# Grammatical gender associations outweigh topical gender bias in crosslinguistic word embeddings

**Katherine McCurdy and Oğuz Serbetçi**
Babbel
{kmccurdy, oserbetci}@babbel.com

## Abstract

Recent research has demonstrated that vector space models of semantics can reflect undesirable biases in human culture. Our investigation of crosslinguistic word embeddings reveals that topical gender bias interacts with, and is surpassed in magnitude by, the effect of grammatical gender associations, and both may be attenuated by corpus lemmatization. This finding has implications for downstream applications such as machine translation.

## 1 Introduction

Gender bias has been demonstrated for English using the widely available Google News embedding (Bolukbasi et al., 2016). For example, a stereotypical view of gender in which career-topic words such as "business" and "executive" are more closely associated with men than women (shown in human subjects by the Implicit Association Test; Nosek et al., 2002) manifests in vector space as systematically shorter cosine distances between embeddings for career words and embeddings for male-associated words (relative to embeddings for female-associated words; Caliskan-Islam et al., 2016). Typologically, English is considered a natural gender language (Stahlberg et al., 2007): the male-female distinction is expressed only on personal pronouns (e.g. he-she), and lexically referenced by a relatively small set of nouns (e.g. husband-wife, actor-actress). In contrast, languages with grammatical gender encode the masculine-feminine distinction as a grammatical feature. Personal nouns typically express a grammatical gender that aligns with the (perceived) sex of the referent (e.g. German: *der Student* the male student, *die Studentin* the female student), and dependent articles and adjectives must agree in gender with the noun (e.g. Spanish: *el niño pequeño* the small boy, *la niña pequeña* the small girl). Grammatical gender is marked on common nouns as well, extending the masculine-feminine distinction to inanimate referents without any apparent semantic basis for category membership. This relative arbitrariness is evident in crosslinguistic discrepancies; for example, in Spanish the sun is masculine (*el sol*) and the moon is feminine (*la luna*), but German has the reverse (*die Sonne*, *der Mond*).

There is a long history of debate on potential social and psychological effects of grammatical gender (Stahlberg et al., 2007). Recently, Boroditsky has presented evidence for cognitive associations between grammatical gender (masculine-feminine) and natural gender (male-female) in object representation for German and Spanish speakers (Boroditsky et al., 2003; Phillips and Boroditsky, 2003; but see Mickan et al., 2014). One oft-cited potential cause is the high frequency of grammatical gender expression: "[t]he sheer weight of repetition (of needing to refer to objects as masculine or feminine) may leave its semantic traces" (Phillips and Boroditsky, 2003, 929). As word embeddings are learned on the basis of distributional characteristics, the "sheer weight of repetition" of grammatical gender marking might profoundly affect the resulting representations.

Our research explores the following questions:

- Do word embeddings in languages with grammatical gender show the same topical semantic biases (e.g. career:male, family:female) as for English?

- Do word embeddings in languages with grammatical gender show additional associative biases to natural gender (e.g. masculine objects:male, feminine objects:female)?

- Can training corpus lemmatization (i.e. gender removal) mitigate these observed biases?

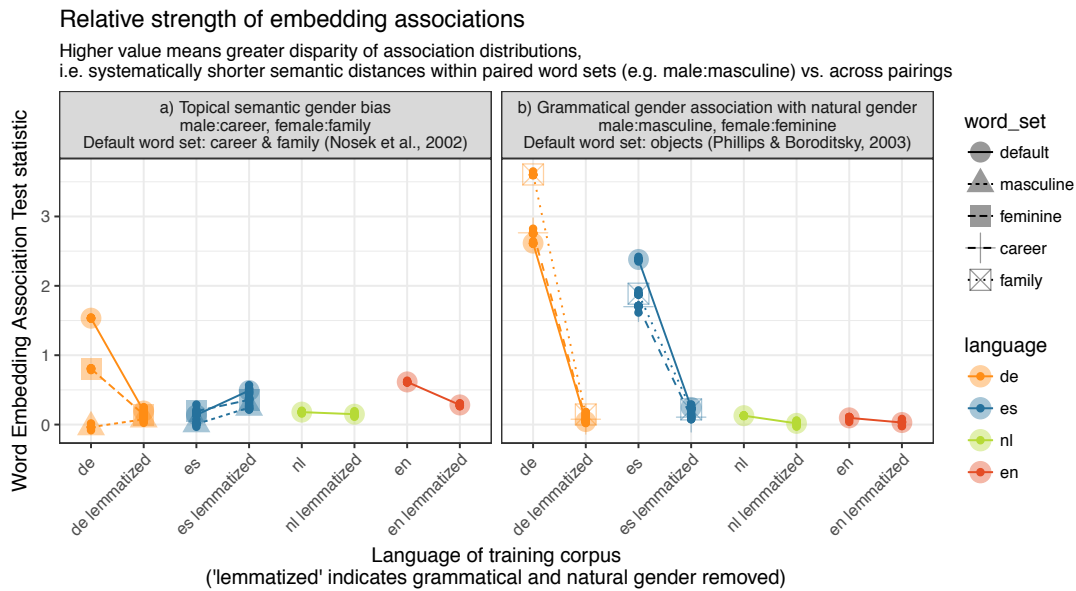

Figure 1: Results of the Word Embedding Association Test.
Left panel (1a) shows topical semantic bias, right (1b) shows grammatical gender associations.

## 2 Method

A parallel corpus was prepared[1] using documents from the OpenSubtitles corpus (Lison and Tiedemann, 2016) with translations in German,[2] Spanish,[3] Dutch,[4] and English,[5] resulting in a corpus of 5,700 documents per language, with total word counts ranging from approximately 2.6 million (Dutch) to approximately 2.9 million (English). A lemmatized corpus was then derived by processing all files using TreeTagger (Schmid, 1994), with additional measures to ensure removal of gender expression (e.g. gendered pronouns in Dutch and English were lemmatized to the common form "he"). For each language and each corpus version (unprocessed and lemmatized), 10 word embeddings were trained using the gensim implementation of the CBOW word2vec model (Řehůřek and Sojka, 2010; Mikolov et al., 2013).

All effects were evaluated using the Word Embedding Association Test methodology developed by Caliskan-Islam et al. (2016). For a given target word (e.g. "office") and two sets of attribute words (e.g. "male, boy, father..." v.s. "female, girl, mother..."), the relative strength of association between the two attributes is determined by calculating the average cosine distance between target word and first attribute set (e.g. "office":"male, boy...") and subtracting the average cosine distance between the target and the second attribute set (e.g. "office":"female, girl..."). The WEAT test statistic is then calculated by subtracting the summed differences for each set of target words; the resulting measurement captures "the differential association of the two sets of target words with the attribute" (Caliskan-Islam et al., 2016, 9). A p-value for each comparison was approximated using Caliskan-Islam et al.'s (2016) permutation test. For each reported result, the mean test statistic (m.t.s.), mean effect size (m.e.s.), and mean p-value (m.p.v.) indicate values averaged over the 10 embeddings for that language and corpus version.

Topical semantic bias was evaluated using the **male:career-female:family** stimuli from the Implicit Association Test (Nosek et al., 2002). Grammatical gender bias was evaluated along the **male:masculine-female:feminine** dimension, where the male-female attribute word sets were the same as used in the IAT, and common object nouns with opposite gender in German and

---

[1] Resources for preparation and analysis: github.com/kmccurdy/w2v-gender

[2] Three categories: masculine, feminine, and neuter.

[3] Two categories: masculine and feminine.

[4] Two grammatical gender categories: common and neuter. As these do not correspond to masculine and feminine, Dutch is considered a natural gender language for our purposes (c.f. Stahlberg et al., 2007).

[5] Natural gender.

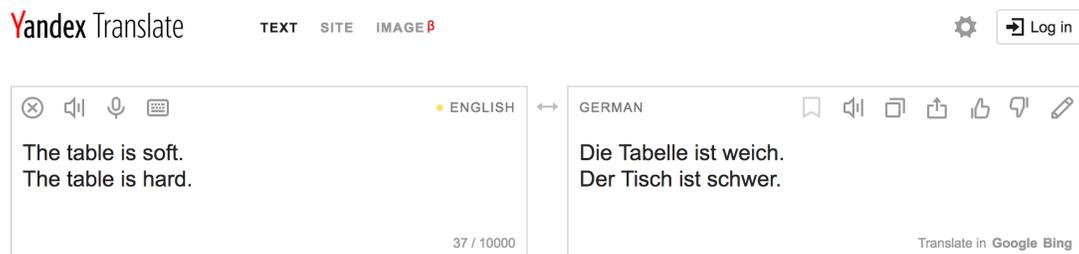

Figure 2: Query to the translation service Yandex.

Spanish (e.g. moon-sun; Phillips and Boroditsky, 2003) comprised the masculine-feminine target word sets. Additionally, we developed a balanced set of target stimuli for German and Spanish for more fine-grained comparison: a set of 5 masculine and 5 feminine nouns each within the topics "career" and "family," yielding a 2x2x2 contrast.

## 3 Results

*Topical semantic gender bias appears reliably in natural gender languages (Fig. 1a).* While all languages show a positive differential association for male:career-female:family stimuli, the Word Embedding Association Test demonstrates reliable effects only for the natural gender languages English (EN mean test statistic 0.61, mean effect size 1.6, mean p-value $<0.01$) and Dutch (NL m.t.s. 0.18, m.e.s. 1.0, m.p.v. 0.03); the difference is not significant for German (DE m.t.s. 1.53, m.e.s. 0.85, m.p.v. 0.08) or Spanish (ES m.t.s. 0.13, m.e.s. 01.0, m.p.v. $>0.3$).

*Topical semantic gender bias interacts with grammatical gender bias (Fig. 1a).* For Spanish and German, the career-family differential vanishes when comparing only masculine nouns (DE m.t.s. -0.03, ES m.t.s. 0.01); however, both show a much higher effect within feminine-gendered words (ES m.t.s. 0.19, DE m.t.s. 0.80, m.e.s. 1.1, m.p.v. 0.04).

*Grammatical gender effects are much larger than topical semantic effects (Fig. 1b).* Whether looking at objects, career-topic, or family-topic words, the male:masculine-female:feminine association is higher in magnitude than all other comparisons (DE m.t.s. 2.62, m.e.s. 1.9, m.p.v. $<0.01$, ES m.t.s. 2.38, m.e.s. 1.9, m.p.v. $<0.001$), including each language's respective career-family bias.

*Lemmatization mitigates grammatical gender effects (Fig. 1b) and reduces some topical bias (Fig. 1a).* All models from lemmatized corpora show masculine-feminine associations of less than .2, and the career-family bias is reduced for English and German.

## 4 Conclusions

Our key finding is that grammatical gender can yield large artefacts in word embeddings; without correction (e.g. lemmatization), downstream semantic applications may see distorted results. These effects are likely subtle and hard to isolate from the myriad factors influencing most natural language applications, but may still produce observable effects. For example, a recent query on the popular translation site Yandex revealed differences in translating the word *table* from English to German (Fig. 2). For the English sentence "the table is hard," *table* is translated as expected to *der Tisch* (masculine, corresponding to the sense "table-object"); however, for the sentence "the table is soft," *table* is rendered as the less frequent word *die Tabelle* (feminine, corresponding to the sense "table of figures"), yielding an erroneous translation. The underlying cause in this case is far from certain, but this is the type of discrepancy which could reflect the distorting effects of grammatical gender in a vector space semantic model; one can imagine other possible outcomes in domains such as search or question answering.

The research presented here demonstrates that grammatical gender expressed on common nouns can yield large effects in word embedding models. The observation that lemmatization mitigates these effects suggests that these artefacts reflect not valid semantic information, but mere distributional properties of function words in the training corpora, and are therefore likely undesirable to practitioners working with such models. One promising direction for future research would be going beyond lemmatization to explore other, more efficient approaches in addressing this issue.

# References

Tolga Bolukbasi, Kai-Wei Chang, James Y. Zou, Venkatesh Saligrama, and Adam T. Kalai. 2016. Man is to computer programmer as woman is to homemaker? Debiasing word embeddings. In *Advances in Neural Information Processing Systems*. pages 4349–4357. http://papers.nips.cc/paper/6228-man-is-to-computer-programmer-as-woman-is-to-homemaker-debiasing-word-embeddings.

Lera Boroditsky, Lauren A Schmidt, and Webb Phillips. 2003. Sex, syntax, and semantics. *Language in mind: Advances in the study of language and thought* pages 61–79.

Aylin Caliskan-Islam, Joanna J. Bryson, and Arvind Narayanan. 2016. Semantics derived automatically from language corpora necessarily contain human biases. *arXiv preprint arXiv:1608.07187* https://arxiv.org/abs/1608.07187.

Pierre Lison and Jörg Tiedemann. 2016. OpenSubtitles2016: Extracting Large Parallel Corpora from Movie and TV Subtitles. In *Proceedings of the 10th International Conference on Language Resources and Evaluation (LREC 2016)*. http://stp.lingfil.uu.se/ joerg/paper/opensubs2016.pdf.

Anne Mickan, Maren Schiefke, and Anatol Stefanowitsch. 2014. Key is a llave is a Schlüssel: A failure to replicate an experiment from Boroditsky et al. 2003. *Yearbook of the German Cognitive Linguistics Association* 2(1). https://doi.org/10.1515/gcla-2014-0004.

Tomas Mikolov, Kai Chen, Greg Corrado, and Jeffrey Dean. 2013. Efficient Estimation of Word Representations in Vector Space. *arXiv:1301.3781 [cs]* ArXiv: 1301.3781. http://arxiv.org/abs/1301.3781.

Brian A. Nosek, Mahzarin Banaji, and Anthony G. Greenwald. 2002. Harvesting implicit group attitudes and beliefs from a demonstration web site. *Group Dynamics: Theory, Research, and Practice* 6(1):101–115. https://doi.org/10.1037//1089-2699.6.1.101.

Webb Phillips and Lera Boroditsky. 2003. Can quirks of grammar affect the way you think? Grammatical gender and object concepts. In *Proceedings of the 25th annual meeting of the Cognitive Science Society*. Citeseer, pages 928–933.

Helmut Schmid. 1994. Probabilistic part-of-speech tagging using decision trees. In *New methods in language processing*. Routledge, page 154.

Dagmar Stahlberg, Friederike Braun, Lisa Irmen, Sabine Sczesny, and K Fiedler. 2007. Representation of the sexes in language. *Social communication* pages 163–187.

Radim Řehůřek and Petr Sojka. 2010. Software Framework for Topic Modelling with Large Corpora. In *Proceedings of the LREC 2010 Workshop on New Challenges for NLP Frameworks*. ELRA, Valletta, Malta, pages 45–50.